\documentclass{article}

\usepackage{arxiv}

\usepackage[utf8]{inputenc} 
\usepackage[T1]{fontenc}    
\usepackage{hyperref}       
\usepackage{url}            
\usepackage{booktabs}       
\usepackage{amsfonts}       
\usepackage{nicefrac}       
\usepackage{microtype}      
\usepackage{lipsum}
\usepackage{graphicx}
\graphicspath{ {./images/} }

\title{V-Seek: Accelerating LLM Reasoning  on Open-hardware Server-class RISC-V Platforms}

\author{
 Javier J. Poveda Rodrigo \\
  DAUIN, Politecnico of Turin, Turin, Italy\\
  \texttt{javier.poveda@polito.it} \\
   \And
 Mohamed Amine Ahmdi \\
  DAUIN, Politecnico of Turin, Turin, Italy\\
   \And
 Alessio Burrello \\
  DAUIN, Politecnico of Turin, Turin, Italy\\
   \And
 Daniele Jahier Pagliari \\
  DAUIN, Politecnico of Turin, Turin, Italy\\
   \And
 Luca Benini \\
  ETHZ, Zurich, Switzerland\\
}

\begin{document}
\maketitle
\begin{abstract}
The recent exponential growth of Large Language Models (LLMs) has relied on GPU-based systems. However, CPUs are emerging as a flexible and lower-cost alternative, especially when targeting inference and reasoning workloads. RISC-V is rapidly gaining traction in this area, given its open and vendor-neutral ISA. However, the RISC-V hardware for LLM workloads and the corresponding software ecosystem are not fully mature and streamlined, given the requirement of domain-specific tuning. This paper aims at filling this gap, focusing on optimizing LLM inference on the Sophon SG2042, the first commercially available many-core RISC-V CPU with vector processing capabilities. 
        On two recent state-of-the-art LLMs optimized for reasoning, DeepSeek R1 Distill Llama 8B and DeepSeek R1 Distill QWEN 14B, we achieve 4.32/2.29 token/s for token generation and 6.54/3.68 token/s for prompt processing, with a speed up of up 2.9$\times$/3.0$\times$ compared to our baseline.
\end{abstract}

\section{Introduction}

Hyperscalers (e.g., AWS) and AI deployment companies (e.g., OpenAI) typically accelerate LLM workloads using GPU clusters or dedicated accelerators such as Tensor Processing Units (TPUs). However, many-core CPU acceleration of LLMs has also been recently explored~\cite{cpu1}, as it provides advantages of lower hardware cost and enhanced flexibility, especially relevant for on-premise and low-latency edge servers.
While existing studies mainly target x86 and ARM, recent many-core chips based on the flexible and open-source RISC-V Instruction Set Architecture (ISA) are relatively unexplored~\cite{intro:performancecharacterisation64coresg2042}. To bridge this gap, this work adapts and optimizes a state-of-the-art LLM inference framework (\texttt{llama.cpp}~\cite{intro:gerganov_llama_cpp_2025}) for the first commodity general-purpose, many-core RISC-V platform (Sophon SG2042~\cite{intro:performancecharacterisation64coresg2042}).%
 On two recent open-source LLMs optimized for reasoning (DeepSeek R1 Distill Llama 8B/QWEN 14B), we show speedups over a baseline \texttt{llama.cpp} implementation of up to 3.0$\times$ in token generation and 2.8$\times$ in prompt processing (i.e., prefill) at 4-bit precision, reaching a throughput of 4.32/2.29 and 6.54/3.68 tok/s, respectively. On vanilla Llama 7B, we achieve 6.63 and 13.07 tok/s for generation and prefill, i.e., a 4.3$\times$/5.5$\times$ speedup w.r.t. the baseline, and 1.65$\times$ better w.r.t. the best-reported results on the SG2042~\cite{results:perfxlab}, while being competitive with CPU-based inference on the incumbent x86 architecture.

%
\vspace{-1em}
\section{Methods}
\begin{figure*}[ht]
  \centering
  \includegraphics[width=.95\linewidth]{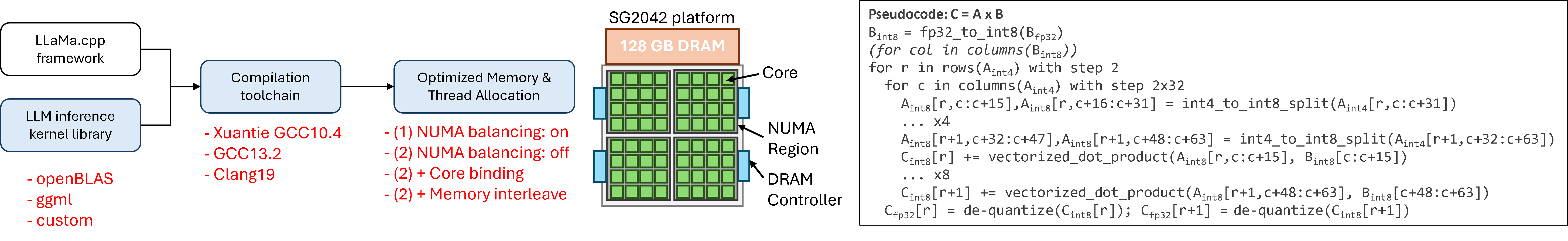}
  \vspace{-0.3cm}
  \caption{From left: optimization flow and contributions. SG2042 block diagram. Pseudocode of the proposed kernel.}
  \vspace{-0.4cm}
  \label{fig:methods_flow}
\end{figure*}
To explore the available alternatives to optimize LLM inference on RISC-V server-class platforms, we target the MILK-V Pioneer, comprising the 64-core Sophon SG2042 and 128GB of DRAM memory.
A block diagram is shown in Fig. \ref{fig:methods_flow}-center.
We identify three directions from which the problem can be attacked, in SW, inspired by works on other architectures~\cite{related:yuan2024llminferenceunveiledsurvey,related:armattention,related:arm_optimization_many}:\\
i) Developing optimized and, if supported, quantized \textbf{kernels} for key LLM layers, that fully exploit the HW, coping with its memory infrastructure, pipeline, and vectorization; 
Fig.\ref{fig:methods_flow}-right, shows the pseudocode of our proposed kernel: first, the \texttt{fp32} input (vector or thin matrix) is quantized to \texttt{int8}; then, two nested loops are executed to perform a GEMV operation, the outermost iterating on the rows of input matrix A, and the innermost on its columns. 
After the column loop ends, de-quantization is applied, combining scale factors from A blocks and B to produce an output \texttt{fp32} value. This new kernel exploits the platform's vector units while also optimizing data locality.\\
ii) Choosing a suitable \textbf{compilation} toolchain, supporting advanced optimization passes and exploiting the available ISA extensions. In our case, kernels are compiled with the Xuantie fork of GCC 10.4, as it is the only one supporting the HW vector units of the Sophon SG2042. Instead, for the whole \texttt{llama.cpp} framework, we consider two alternatives: GCC 13.2, and Clang 19 (Xuantie GCC 10.4 is not compatible with the latest \texttt{llama.cpp} release).\\
iii) Optimizing model \textbf{mapping}, specifically pages/thread allocation, addressing this type of system's complex memory hierarchy. Namely, we optimize Non-uniform Memory Access (NUMA) latency exploring different \textit{numactl} options combined in 4 policies: i) NUMA Balancing on, all other options off, ii) all options off, iii) Balancing off+Core Binding on, iv) Balancing off+Memory Interleaving on.

We apply our optimizatons to the \texttt{llama.cpp} \cite{intro:gerganov_llama_cpp_2025} framework, testing on 3 open-source LLMs of increasing size, with Q4\_0 quantization (vanilla Llama 7B, DeepSeek R1 Distill Llama 8B, DeepSeek R1 Distill QWEN 14B, referred to as 7B, 8B and 14B below).

\vspace{-1em}
\section{Results}
To show the results of our optimization, we executed the prefill of the three LLMs with user prompt "Explain to me what is RISC-V, what are its principles and why it is so cool?" (22 tokens), while we averaged the token generation performance over 256 test-generated tokens.
\begin{figure}[t]
  \centering
  \includegraphics[width=0.6\linewidth]{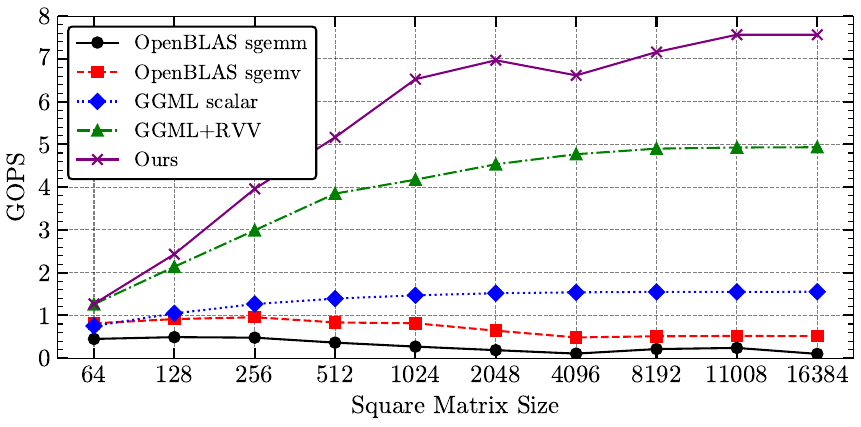}
  \vspace{-0.2cm}
  \caption{Matrix vector multiplication size scalability test}
  \label{fig:gemv_bench}
  \vspace{-0.6cm}
\end{figure}
\textbf{Kernel Scaling.} Fig.\ref{fig:gemv_bench} shows the single-thread scalability of multiple baseline kernels (llama.cpp's GGML and OpenBLAS's defaults) and of our proposed one.
Compared to the best baseline, we improve the GOPS by $+38.3\%$ on average, peaking at $+56.3\%$ at matrix size 1024.

\begin{figure}[ht]
  \centering
  \includegraphics[width=0.6\linewidth]{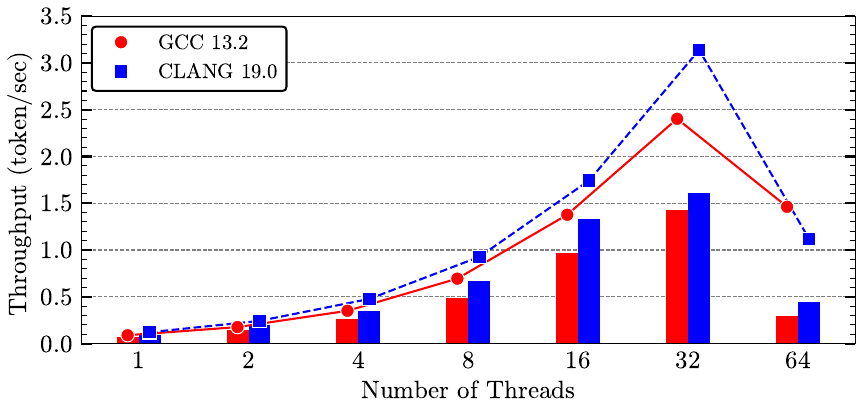}
  \vspace{-0.2cm}
      \caption{Compilers comparison scaling the n. of threads for DeepSeek's 8B model token gen., Bar, and prefill, Line.}
  \vspace{-0.3cm}
  \label{fig:scalability_comp}
\end{figure}
\textbf{Compiler exploration.} In Fig.\ref{fig:scalability_comp}, we evaluate DeepSeek's 8B model inference when compiling with Clang or GCC, using our proposed kernel.
Clang 19 constantly outperforms GCC 13.2, reaching average performance gains of $34\%$ and $25\%$ for token generation and prefill, respectively.
The crucial reason is the combination of ISA extension support, and more advanced compilation passes (e.g., more aggressive in-lining and loop unrolling).
Regardless of the compiler used, using $>32$ threads leads to a performance loss. This behavior is attributed to the default NUMA balancing policy, which is suboptimal for LLM inference due to the predictable nature of the workload, leading to a high number of thread and memory page migrations.

\begin{figure}[ht]
  \centering
  \includegraphics[width=0.6\linewidth]{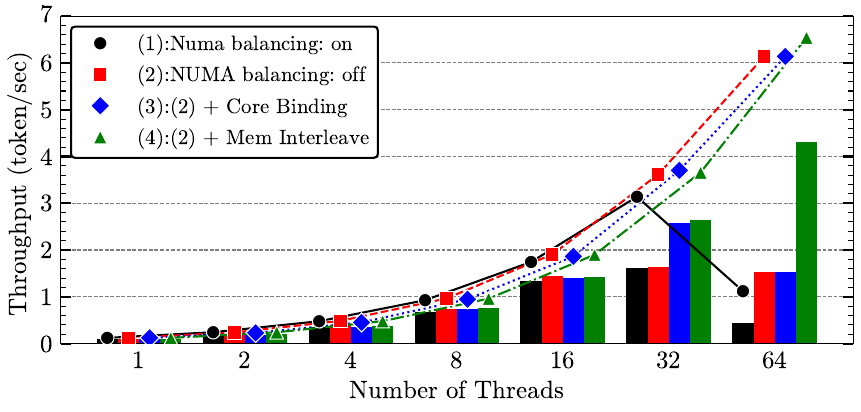}
  \vspace{-0.2cm}
  \caption{NUMA policies exploration on DeepSeek's 8B model. Token generation shown with bars, prefill with lines.}
  \label{fig:numa_comparison}
  \vspace{-0.3cm}
\end{figure}
\textbf{NUMA policy impact.} Indeed, with the NUMA balancing off and memory interleaving on, as expected, we achieve the best results for both token generation (4.32 tokens/s) and for prefill (6.54 tokens/s) with 64 threads, thanks to the strong reduction in memory page migration.

Overall, thanks to our optimizations, the 7B, 8B and 14B LLMs reach a maximum throughput of 13.07/6.54/3.68 tok/s respectively, outperforming a baseline \texttt{llama.cpp} by up to 5.5$\times$/2.9$\times$/3$\times$. Compared to the best reported result on the SG2042~\cite{results:perfxlab}, we improve the peak throughput on LLama 7B by 1.65$\times$. Versus a similar and more mature x86 platform, the 64-cores AMD EPYC 7742, we improve the energy efficiency by $1.2\times$(55 token/s/mW vs 45  token/s/mW)~\cite{result:QIGen}.

\vspace{-1em}

\end{document}